\documentclass[10pt,twocolumn,letterpaper]{article}

\usepackage{cvpr}              

\definecolor{cvprblue}{rgb}{0.21,0.49,0.74}
\usepackage[pagebackref,breaklinks,colorlinks,allcolors=cvprblue]{hyperref}
\usepackage{amsmath,graphicx,multirow,booktabs,makecell}

\def\paperID{8386} 
\def\confName{CVPR}
\def\confYear{2025}

\title{Towards Understanding How Knowledge Evolves in Large Vision-Language Models}

\author{Sudong Wang$^{1,2}$, Yunjian Zhang$^3$\footnotemark[1], Yao Zhu$^3$\footnotemark[1], Jianing Li$^3$\\
Zizhe Wang$^3$, Yanwei Liu$^1$, Xiangyang Ji$^3$\\
$^1$Institute of Information Engeering, Chinese Academy of Sciences; \\
$^2$Nanyang Technological University; $^3$Tsinghua University\\
\texttt{\small SWANG049@e.ntu.edu.sg, sdtczyj@gmail.com, ee\_zhuy@zju.edu.cn}
}

\begin{document}
\maketitle

\footnotetext[1]{Corresponding authors: Yunjian Zhang and Yao Zhu.}

\begin{abstract}
Large Vision-Language Models (LVLMs) are gradually becoming the foundation for many artificial intelligence applications. However, understanding their internal working mechanisms has continued to puzzle researchers, which in turn limits the further enhancement of their capabilities. In this paper, we seek to investigate how multimodal knowledge evolves and eventually induces natural languages in LVLMs. We design a series of novel strategies for analyzing internal knowledge within LVLMs, and delve into the evolution of multimodal knowledge from three levels, including single token probabilities, token probability distributions, and feature encodings. In this process,  we identify two key nodes in knowledge evolution: the critical layers and the mutation layers, dividing the evolution process into three stages: rapid evolution, stabilization, and mutation. Our research is the first to reveal the trajectory of knowledge evolution in LVLMs, providing a fresh perspective for understanding their underlying mechanisms. Our codes are avaiable at \url{https://github.com/XIAO4579/Vlm-interpretability}.
\end{abstract}    
\section{Introduction}
\label{sec:intro}

Benefiting from the rapid progress of Large Language Models (LLMs) \cite{brown2020language,vicuna2023,driess2023palm,gilardi2023chatgpt,rohan2023alpaca,touvron2023llama}, Large Vision-Language Models (LVLMs) \cite{bai2023qwen,dai2023instructblip,zhang2023internlm,brooks2023instructpix2pix,black2023training,li2023llavamed,zhu2023minigpt,zhang2023gpt4roi,liu2023llava,liu2023improvedllava,ye2023mplug,He2024malmm,Zhang2024groundhog,Chen2023internvl,Yuan2023osprey,Dong2024dreamllm,Cha2023honeybee} have made remarkable strides in recent years. These models align multimodal features in the same latent space and demonstrate exceptional performance in jointly representing complex visual and linguistic patterns, elevating general foundational models to unprecedented levels.

In a typical processing flow of an LVLM, visual and textual inputs are first encoded into vectors by their corresponding feature extractors, respectively. The multimodal features are then aligned through an affine module and embedded into a format suitable for the language model. Ultimately, the embedded features are fed into the backend LLM to generate natural language descriptions. This raises a critical question: \textbf{How are multimodal features converted into natural language?} Exploring this question is vital for demystifying the black-box LVLMs and is essential for developing more intelligent models.

Currently, several studies have explored the knowledge transfer process within LLMs \cite{chuang2023dola,voita2019bottom,meng2022locating,chen2023beyond,Dai2021knowledge,chefer2021transformer,Durrani2020analyzing}, analyzing the evolution of representations across different layers from a bottom-up perspective. 
Although these studies provide valuable insights into the processing mechanisms of language models, research on bridging visual and linguistic representations within LVLMs is still limited.

\begin{figure}[ht]
	\centering
	\includegraphics[width=.48\textwidth]{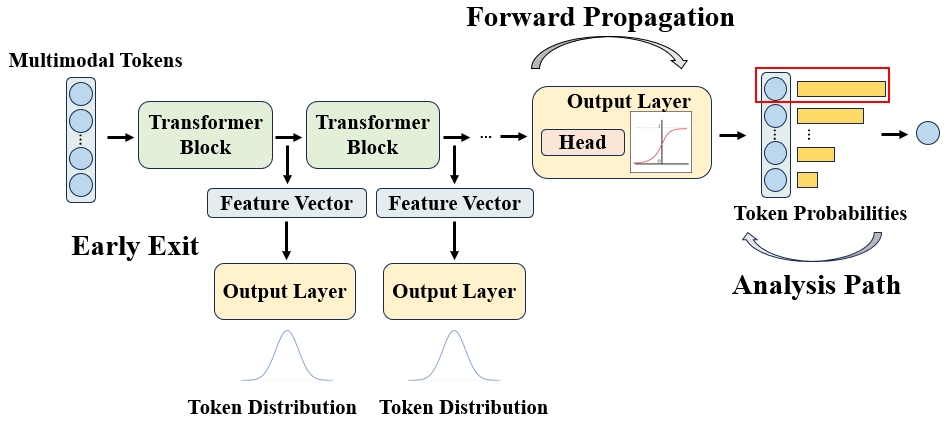}
	\caption{The analysis path in this paper. We begin with the probabilities of single tokens, then delve into the probability distributions of all tokens over the vocabulary set, and finally analyze the hidden features used for generating the tokens.}
	\label{fig:path}
\end{figure}

In this work, we investigate how aligned multimodal features are transmitted within LVLMs and ultimately converted into natural languages, aiming for finding the evolution patterns of knowledge representations, and bridging the gap between multimodal information and natural language. Towards this end, we design a series of effective methods for tracking the knowledge flow across intermediate layers, without any additional special training, these methods allow us to intuitively observe how knowledge evolves through the hidden space.
As shown in Figure \ref{fig:path}, we reverse-engineer the token generation procedure for investigation. Specifically, we first examine the probability changes of key tokens across layers, then study the differences in token probability distributions between layers, and finally delve into the feature encoding level to explore the progression of knowledge from shallow to deep layers. This approach enables us to start from the output and backward reason how language is constructed on multimodal features.

At the \textbf{token probability level}, we use the early exit technique \cite{elbayad2020depth,schuster2022confident,mahsereci2017early,zhou2020bert,matsubara2022split,ghodrati2021frameexit} to calculate the probabilities of predicted tokens across different transformer blocks. 
We observe that \emph{token probabilities undergo an abrupt shift in the middle layers, jumping from near-zero to a significantly higher value, placing them in the candidate list of the output tokens. Additionally, some tokens experience a second mutation in probability near the output layers, causing the dominant token to change.} Furthermore, we demonstrate the close correlation between this mutation phenomenon and model hallucinations in the following experiments, providing a new perspective for uncovering the causes of hallucinations in LVLMs.

At the \textbf{probability distribution level}, we compute the probabilities of all tokens in the vocabulary set at each layer, then calculate the Jensen-Shannon (JS) divergence of the distributions between adjacent layers, implicitly revealing the knowledge variation across layers. We observe \emph{a distinct hierarchical phenomenon in JS divergences of the probability distributions between adjacent layers. In the shallow layers, the JS divergence values are relatively high, indicating significant differences in probability distributions. 
As we move into the middle layers of the network, the JS divergence values decrease sharply, and the probability distributions of the layers begin to converge.} This phenomenon marks the evolution of multimodal knowledge in the model from a rapid evolution phase to a stable phase. Further investigations have also demonstrated that this change of evolution stage holds significant potential for achieving the compression of model depth.
Another intriguing phenomenon is that \emph{the probability distributions of some tokens may experience mutations after convergence, leading to shifts in knowledge representation that ultimately affects the selection of output tokens.}

We delve deeper into the \textbf{feature encoding level}, applying t-SNE on the encoding vectors used to predict tokens in the hidden space, compressing the high-dimensional features into 2D vectors. It allows us to examine conveniently the trend of feature changes throughout the next-token prediction process. We observe that \emph{the feature vectors of all tokens exhibit a tight aggregation at the initial layer, and as depth increases, the features in subsequent layers radiate outward in an approximately linear pattern.} This finding highlights the compactness and continuity of knowledge evolution in LVLMs at the feature level, providing valuable insights for understanding LVLM generalization and for designing more efficient finetuning methods. 

It is important to note that although our experiments are conducted on the language model of LVLMs, the phenomena described above are unique to LVLMs, and have not been observed in LLMs with pure text inputs. These phenomena outline the knowledge evolution process in LVLMs: in the shallow layers, multimodal knowledge evolves rapidly, gradually exhibiting token-specific characteristics; starting from the middle layers, the evolution speed significantly slows down, with a certain degree of similarity in the knowledge acquired by each layer; for some tokens, when knowledge is transferred to deep layers, it often undergoes a mutation, leading to a secondary evolution, the outcome of which profoundly influences the model's final prediction.

Our main contributions are summarized below:
\begin{itemize}
	\item We deconstruct the token generation process in LVLMs and, for the first time, investigate deeply the evolution of multimodal knowledge within the model in multiple levels, thereby building a bridge between multimodal inputs and natural language descriptions.
	\item We design various effective strategies to analyze the evolution at different levels, and observe several novel and intriguing phenomena from extensive experiments.
	\item We creatively reveal the patterns of knowledge evolution in LVLMs, which has great significance for illuminating the mechanisms underlying the black-box models and designing more efficient and accurate LVLMs.
\end{itemize}

\section{Backgrounds}
\label{sec:pre}

Generally, an LVLM consists of three modules \cite{Cai2023vipllava,Jin2023chatunivi,Xu2024Libra,Li2023monkey,Zhu2024languagebind,Rasheed2023glamm,Tang2024salmonn,Sun2023generative}. The first module is the modality encoder, which generally includes a visual encoder and a text encoder, extracting features from the input modalities. Pretrained encoders are usually used here, such as CLIP \cite{radford2021learning} and BLIP \cite{li2022blip}. Next is the feature alignment module, ensuring effective interaction between multimodal inputs in the same feature space. 
The final module is an LLM, using the fused features to generate natural language descriptions in an autoregressive manner, progressively predicting the next word. In LVLMs, pretrained open-source LLMs, such as LLaMA \cite{touvron2023llama} and Vicuna \cite{vicuna2023}, are typically used, which are then finetuned with visual-language instructions.

Apart from the pretrained encoders and LLMs, the primary difference among various LVLMs lies in the feature alignment methods, mainly categorized into two types: linear projections that directly transform image features into the word embedding space \cite{liu2023llava,lin2024vila}, and sampling methods that learn to represent visual-textual alignments \cite{ye2023mplug,dai2023instructblip,zhu2023minigpt}. Regardless of the method used by the alignment module, the fused multimodal features do not affect the subsequent language generation process. To ensure the generalizability of our study, we do not delve into the intrinsic workings of the alignment module but instead focus on the process of converting visual-textual features to languages.

We conduct analyses on the LLaVA-1.5 model \cite{liu2023llava}, with a 32-layer vicuna-1.5 model \cite{vicuna2023} as its backend language module. The aligned visual tokens are denoted as $x_v=\{x_0,x_1,...,x_{M-1}\}$, where $M$ is the length of the visual tokens and it is usually a fixed number. The input text is also tokenized into a set of tokens, denoted as $x_t=\{x_M,x_M,...,x_{M+N-1}\}$, where $N$ is the length of the text tokens, which depends on the user prompt. The LLM predicts tokens in an autoregressive manner based on previous tokens, assuming $K$ tokens have been predicted, they can be denoted as $x_p=\{x^o_0,x^o_1,...,x^o_{K-1}\}$. Therefore, the final input sequence of the LLM can be denoted as $x_f=\{x^v,x^t,x_p\}$, whose length is $M+N+K$.

When predicting the ($K$+1)-th token,  the input is first embedded into a sequence of vectors $H^0=\{h_0^0,h_1^0,...,h_K^0\}$. After that, $H^0$ is processed by $L$ stacked transformer blocks successively (for LLaVA-1.5-7B, $L=32$), then the language head $\phi(\cdot)$ is applied on the output of the final layer ($H^L$) to predict the probability of the next token over the vocabulary set $\chi$, formulated as
\begin{equation}
	p(x^o_{K}|x_p)=\text{softmax}(\phi(H^L))_{x^o_K},\text{ } x^o_K\in\chi.
\end{equation}

After obtaining all the logits of tokens in $\chi$, the predicted token is selected by a decoding method, then the selected token is concatenated to the last of $x_p$ for the next-round generation, until all tokens are predicted.

To analyze the inner knowledge representations of the model, we apply the function $\phi(\cdot)$ on each intermediate layer, and predict next tokens with these hidden features. Denoting the output of the $j$-th layer as $H^j$, the probability of the next token predicted with it can be calculated as
\begin{equation}
	\begin{aligned}
		&p_j(x^o_{K}|x_p)=\text{softmax}(\phi(H^j))_{x^o_K}\\
		&x^o_K\in\chi,\text{ } j\in\{0,...,L-1\}
	\end{aligned}.
\end{equation}

Our study starts from the probabilities of the output tokens in the intermediate layers, and then analyzes the probability distribution of all tokens over the vocabulary set, and finally, explores further the feature vectors that used for calculate token probabilities.
\begin{figure*}[ht]
	\centering
	\includegraphics[width=1.\textwidth]{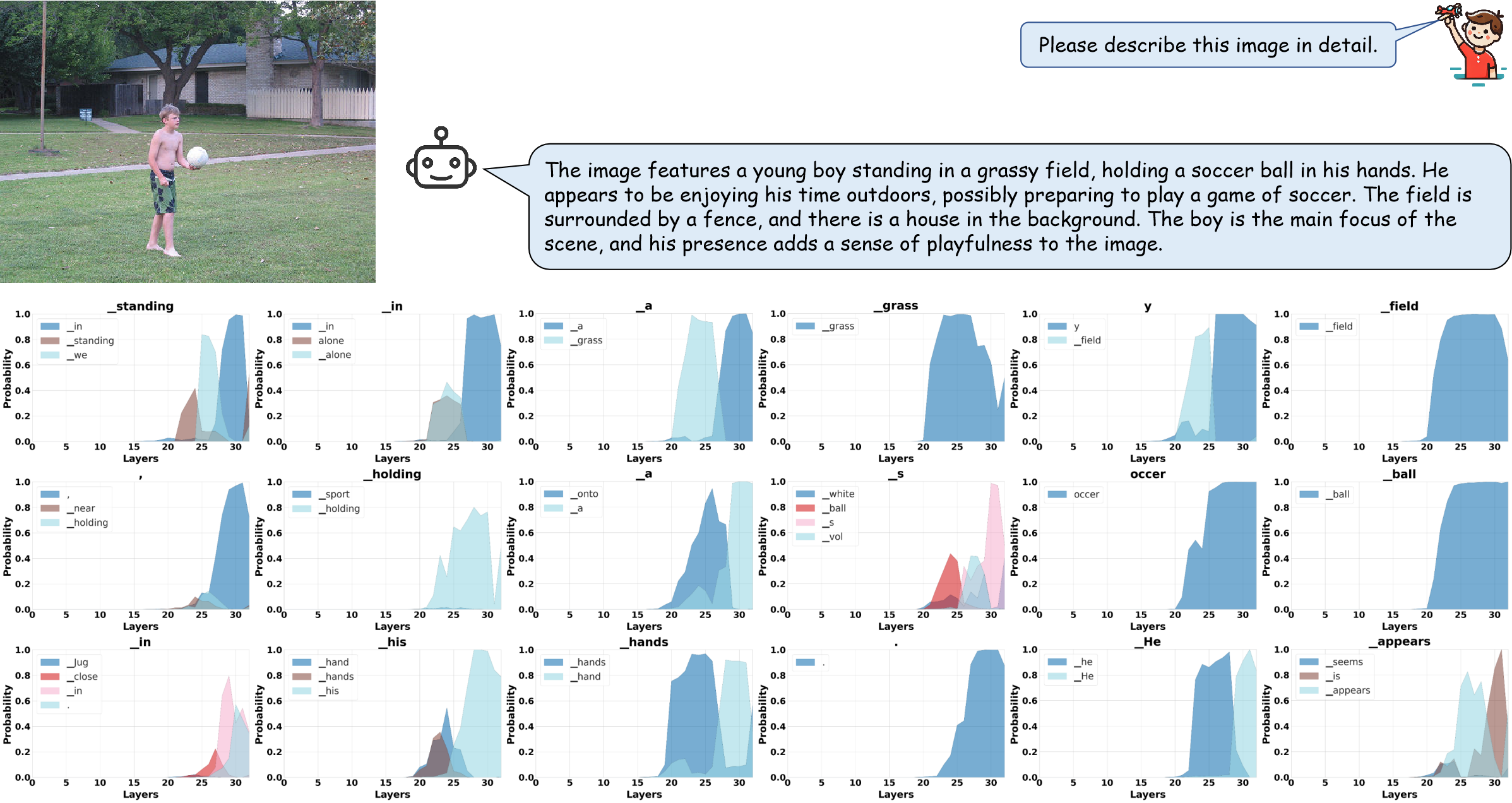}
	\caption{The probabilities of tokens across different layers during normal inference processes.}
	\label{fig:correcttoken}
\end{figure*}
\section{Token Probability Analyses}
\label{sec:token}

In this section, we analyze how the probability of individual tokens varies across different layers. We first investigate typical token probability trends during normal inferences, followed by an examination of probability changes for key entities when the model generates hallucinations.

\subsection{Normal Inference}
Our analyses are performed on the AMBER \cite{wang2023llm} and MSCOCO datasets \cite{lin2015coco}. We use ``Please describe this image in detail'' to prompt LLaVA-1.5 to generate descriptions, and apply early exit to compute the probabilities for each token across different layers. Taking the AMBER-024 image as an example, we illustrate our findings, and the results are presented in Figure \ref{fig:correcttoken}, in which 18 tokens in the generated text are shown. The observations are as follows:

First, all tokens exhibit very low probabilities in the shallow layers of the model, nearing zero. However, around the 20-th layer, the probabilities of these token sharply increase (for ease of description, we refer to these layers as critical layers), standing out from the vocabulary set to gain a probabilistic advantage. Some tokens maintain this advantage through to the output layer, ultimately being included in the generated description text.

Second, for some tokens that are easy to predict, such as punctuation marks, words directly copied from the input question like ``image'', and those with high certainty due to obvious features, their probabilities tend to stabilize after the critical layers. These tokens exhibit minimal variations and maintain high probabilities up to the output layer. In contrast, when predicting tokens that contain important information—such as certain verbs, nouns, and prepositions describing object attributes (e.g., ``standing'', ``in'', ``boy'', ``he'', ``appear'')—the number of candidate tokens increases, and these tokens often have high probabilities. Additionally, both the predicted tokens and other candidates experience abrupt changes in probability, with the predicted token only gaining an advantage after this probability surge, eventually becoming the final selection.

\begin{figure*}[ht]
	\centering
	\includegraphics[width=1.\textwidth]{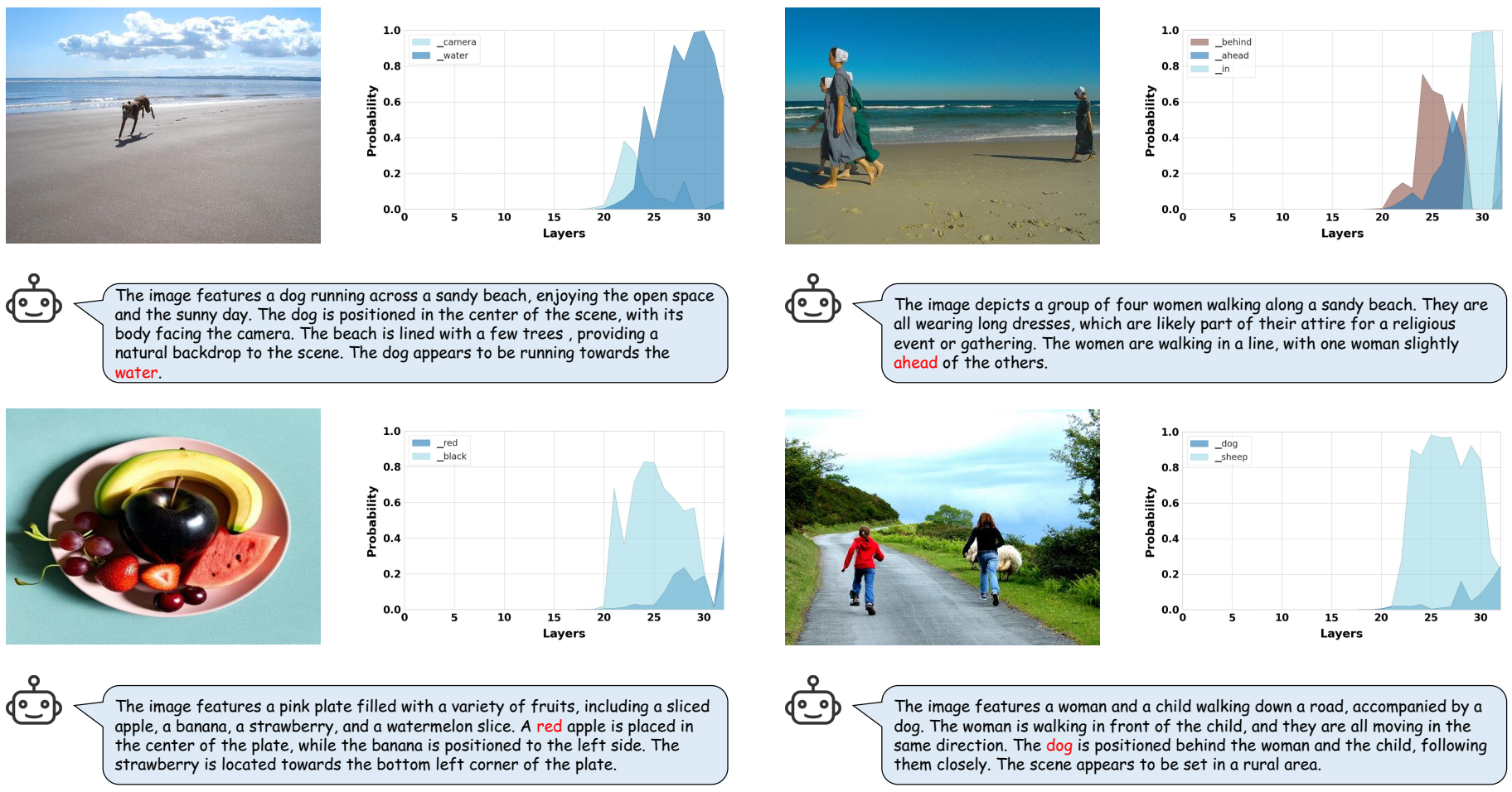}
	\caption{The probabilities of hallucinated tokens and correct tokens across different layers when the LVLM falls into hallucinations.}
	\label{fig:hallutoken}
\end{figure*}

\subsection{Hallucinations}
The probability mutation in some entities is an intriguing phenomenon. To further analyze its impact, we examine the probability changes of hallucinated tokens when the model generates faithless descriptions, and several examples are placed in Figure \ref{fig:hallutoken}. It can be observed that all hallucinated tokens, including ``water'', ``ahead'', ``red'', and ``dog'', experience probability mutations after the critical layers, and the layers where the probabilities mutate are referred as the mutation layers. Most of these tokens do not exhibit probability advantages before the mutation layers; instead, the correct tokens, such as ``camera'', ``behind'', ``black'', and ``sheep'', often have higher probabilities. After the mutation layers, the probabilities of correct tokens rapidly decline, while those of hallucinated ones rise sharply, enabling them to become the dominant tokens and ultimately be output. This phenomenon suggests that probability mutations in deep layers may be associated with hallucinations. One possible reason is that, in LVLMs, the language model's capability—significantly large in both parameters and training data—far surpasses that of the visual model \cite{leng2023mitigating,wang2024mitigating}, resulting in that the language generation process is dominated by the language model. Once the knowledge from the visual model stabilizes, it becomes difficult for subsequent layers to acquire new knowledge from the input. As a result, the language model may inject its own prior knowledge to continue driving the evolution of knowledge, such as ``apples are red'' or ``a dog’s tail is fluffy''. For the visual module, this constitutes external knowledge, as it does not originate from the image itself. When such knowledge conflicts with visual features, hallucinations may arise.

\begin{figure*}[ht]
	\centering
	\includegraphics[width=1.\textwidth]{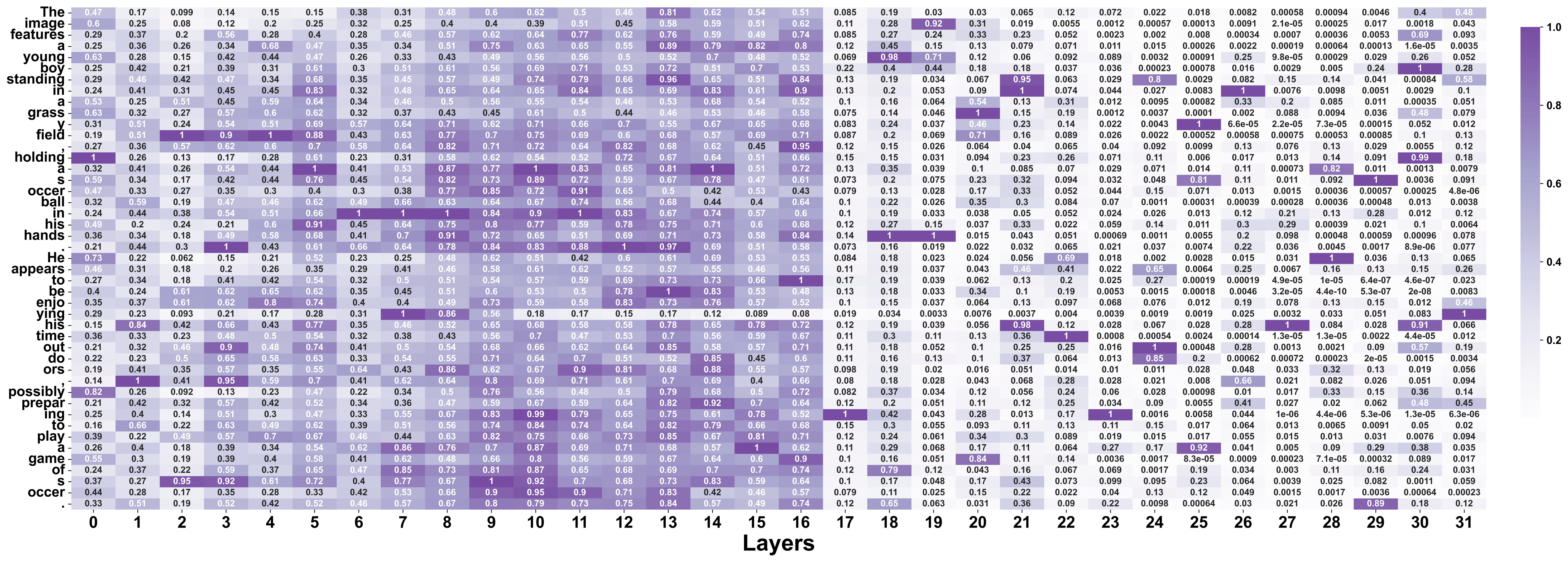}
	\caption{The JS divergences of token probability distributions across adjacent layers during normal inference processes.}
	\label{fig:distribution}
\end{figure*}

\section{Token Distribution Analyses}
\label{sec:dis}

To further explore the evolution of knowledge and understand the reason for the behavior of single token probabilities, we analyze the probability distribution of all tokens in the vocabulary set across each layer and calculate the Jensen-Shannon (JS) divergence between adjacent layers, which can be formulated as
\begin{equation}
	\text{JSD}(p_i \parallel p_j) = \frac{1}{2}( \text{KLD}(p_i \parallel A)+\text{KLD}(p_j \parallel A)),
\end{equation}
\begin{equation}
	\text{KLD}(p_i \parallel A) = \sum_{k} p_i(k) \log \left( \frac{p_i(k)}{A(k)} \right),
\end{equation}
where $i$ and $j$ are two adjacent layers, $A$ is the average distribution of $p_i$ and $p_j$.

\subsection{Normal Inference}
The analyses are also performed on the AMBER and MSCOCO dataset, and we still take the image in Figure \ref{fig:correcttoken} as an example to illustrate our findings, which is shown in Figure \ref{fig:distribution}. Two patterns can be observed from Figure \ref{fig:distribution}. \textbf{\#1:} Around the 18-th layer, there is a clear distinction in the JS divergence values between the shallow and deep layers. In the shallow layers, the divergence values are large, indicating substantial differences in token distributions between adjacent layers. After the 18-th layer, the divergences rapidly decrease to near zero and remain stable up to the output layer, suggesting similar token distributions across these deep layers. This phenomenon aligns with the first observation in Section \ref{sec:token}, where the critical layers of divergences correspond to the critical layers of token probabilities approximately. This pattern subtly suggests that the process of knowledge evolution unfolds in distinct stages. Large divergences in shallow layers indicate that knowledge evolves rapidly across these layers, causing dramatic shifts in token probability distributions. As shallow layers primarily focus on the local features, it is difficult for a token to gain a probability advantage at this stage. Instead, many tokens may have similar probabilities, thus their probabilities remain near zero. When knowledge evolves into deeper layers, progressing from low-level to high-level feature aggregation, small divergences values mean that the pace of evolution obviously slow down. Consequently, tokens associated with deep knowledge start to gain probability advantages and emerge from the vocabulary set. \textbf{\#2:} For some tokens, especially those containing important information, such as ``boy'' and ``standing'', the divergence values in deep layers often mutate, which indicates that token distributions undergo significant changes in these layers, further suggesting that the knowledge within the model also experiences sudden shifts. This pattern can explain the second observation in Section \ref{sec:token}. In the mutation layers, new knowledge is injected into the model, causing a secondary evolution of the knowledge and leading to oscillations in the token probability distributions. As a result, the probability of individual tokens related to the knowledge also undergoes sudden changes. If the injected knowledge conflicts with the image-derived knowledge already present in the model, it may lead to the emergence of a new dominant token, potentially causing the model to fall into hallucinations.

\begin{figure*}[ht]
	\centering
	\includegraphics[width=.95\textwidth]{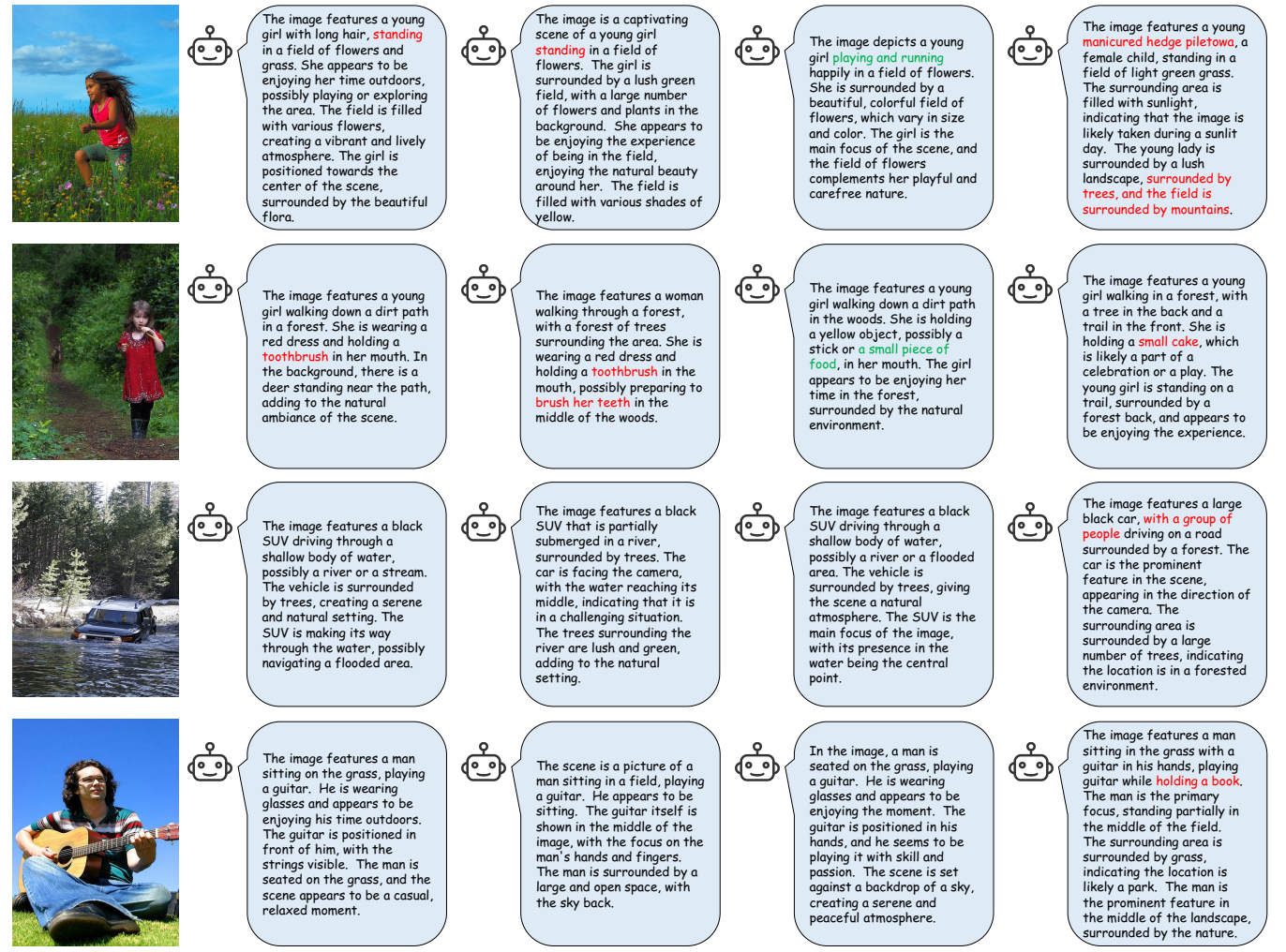}
	\caption{The effect of skip connections on model's output. From left to right: the original image, the descriptions from the original model, the descriptions when skipping from the critical layers to the mutation layers, the descriptions when only skipping the mutation layers, and the descriptions when skipping from the critical layers to the last few layer (as the layers near the output contain linguistic priors, we retain the final 5 layers). Hallucinated tokens are marked in red, and corrected tokens are marked in green.}
	\label{fig:skip}
\end{figure*}

\subsection{Skip Connection}
According to Figure \ref{fig:distribution}, the critical layers and mutation layers divide knowledge evolution into three stages: rapid evolution, stabilization (slow evolution), and mutation. In this experiment, we explore the impact of the latter two stages through skip connections: jumping from the critical layers to the mutation layers (\textbf{skip.1}), skipping only the mutation layers (\textbf{skip.2}), and jumping directly from the critical layers to the last few layers (\textbf{skip.3}). Some examples of the experimental results are shown in Figure \ref{fig:skip}.

We observe when skipping the layers before the mutation layers, the model's output remains semantically close to the original, though some wording may differ slightly, and even the hallucinations in the original descriptions appear in the \textbf{skip.1} output. This indicates that during the phase from the critical layers to the mutation layers, knowledge changes minimally between layers, which means the evolution process of multimodal knowledge tends to stabilize. This finding may be valuable for helping researchers design lightweight algorithms to compress model's depth. In the \textbf{skip.2} setup, the model preserves most of the original semantics (especially for the words before the hallucinations) while also correcting some hallucinations, for example, changing ``standing'' to ``playing'' and ``toothbrush'' to ``food''. This suggests that the knowledge injected into the mutation layers primarily influences the expression of certain semantic elements, while leaving most of the overall meaning unchanged. If the injected knowledge misaligns with image features, it may leads to hallucinations. Therefore, handling the mutation layers has the potential to enhance the model's accuracy. When jumping directly from the critical layers to the last few layers, the model's output retains little similarity to the original results and exhibiting more hallucinations, indicating that knowledge continues to evolve after the critical layers, though at a slower pace, potentially involving processes of local knowledge aggregation and global knowledge construction. Therefore, skipping too many layers may cause some knowledge to be omitted, consequently affecting the outputs.
\section{Feature Encoding Analyses}
\label{sec:feature}

This section moves deeper into the feature level, investigating the variation patterns of feature encodings in the process of knowledge evolution. Considering that the features are typically high-dimensional, for more intuitive analysis, we use t-SNE to reduce their dimensions.

\begin{figure}[htbp]
	\centering
	\includegraphics[width=.45\textwidth]{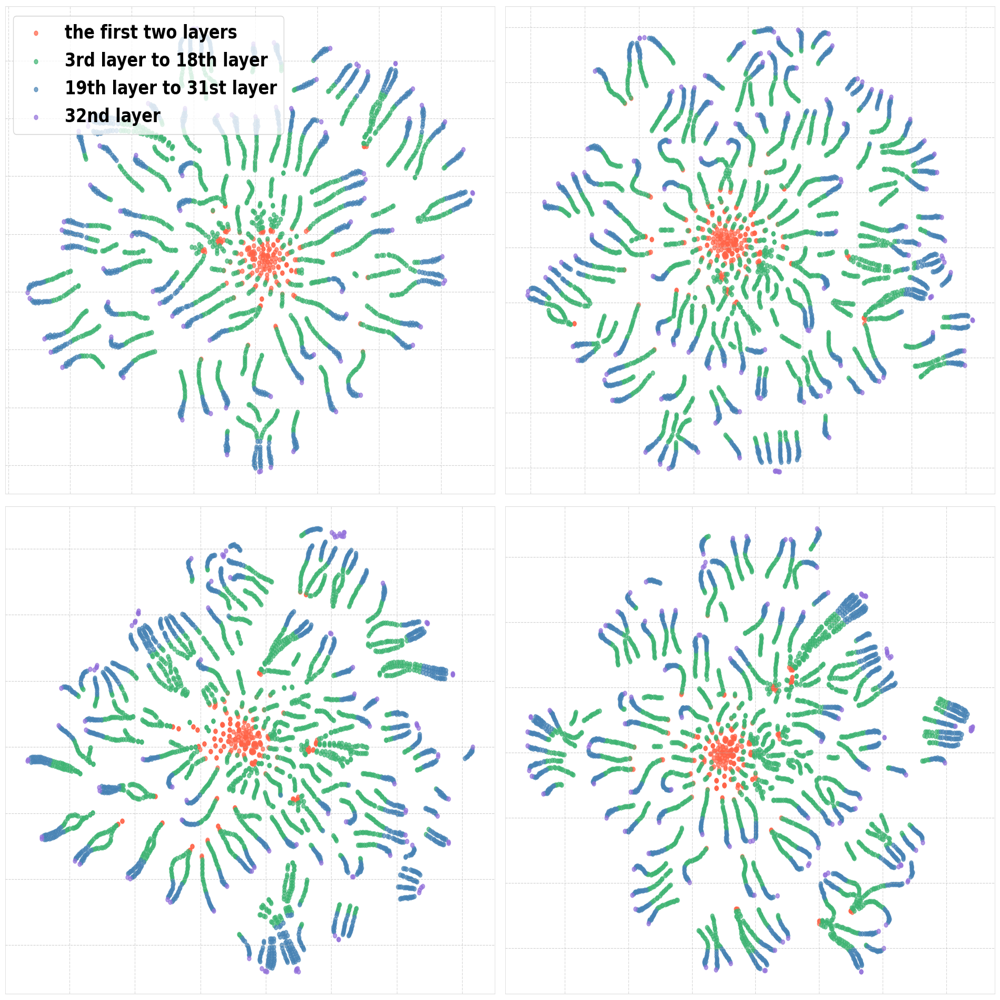}
	\caption{Feature encodings across layers in the same image for different tokens.}
	\label{fig:single}
\end{figure}

\subsection{Features in a Single Image}
We first observe the features when predicting different tokens within a single image. For LLaVA-1.5 with 32 layers, each layer's feature vector has a dimension of 1$\times$4096. Assuming the number of predicted tokens is $T$, the size of the feature matrix to be analyzed is $T\times32\times4096$. Then the compressed matrix after t-SNE is
\begin{equation}
	F^{T\times32\times1}=\text{t-SNE}(F^{T\times32\times4096}).
\end{equation}

We take the four images shown in Figure \ref{fig:hallutoken} as examples, and visualizes their feature encodings in Figure \ref{fig:single}. There are two patterns can be observed. \textbf{\#1:} For all predicted tokens, their feature encodings exhibit a progressively diverging trend. Specifically, the features of all tokens are closely clustered together in the first two layers, indicating that in the initial stage, the features for different tokens are similar. As the depth increases, a clear divergence emerges, and by the 32-nd layer—the final layer of the model—all token features are widely dispersed. This suggests that as information propagates forward from shallow to deep layers, knowledge continues to evolve, gradually revealing token-specific characteristics. \textbf{\#2:} An approximately linear distribution emerges for the feature encodings of a single token across different layers. Starting from the initial layer's feature, the features in subsequent layers diverge gradually along a near-linear path, with each layer's feature located just next to the previous one. Additionally, for some tokens, a hierarchical phenomenon is observed (due to the large number of tokens in the figure, we do not label them to avoid making the figure overly cluttered); for example, some clusters of green points (shallow features) are separate from clusters of blue points (deep features). This indicates that the deep features and shallow features for these tokens differ significantly. Although these features are separated, within each cluster, the features across different layers still exhibit an approximately linear arrangement.

\subsection{Features in Different Images}
We further explore the knowledge evolution patterns with different input images. In this experiment, it is essential that the predicted tokens carry important information, accurately indicating key features of the input images. For two images, $\text{Img}_1$ and $\text{Img}_2$, if the predicted token for $\text{Img}_1$ is a functional word such as ``a'' or ``an'', while the token for $\text{Img}_2$ corresponds to a key object in the image, the knowledge required for these two types of tokens differs significantly. The former primarily depends on contextual relationships and linguistic priors, whereas the latter relies entirely on the input image. Therefore it is unconvincing to observe the evolution patterns by comparing them. 

To make the output tokens more controllable and concentrated on the image contents, we perform this experiment on the visual question-answering (VQA) task. In our setting, the model is prompted to answer simple questions about the input image with a single word, such as ``What is the color of the flower?'' or ``What is the main object of the image?''. This allows us to focus solely on the feature changes across layers by observing how the model outputs the first token (also the only token). Images with the same question are grouped together, with a total count of $N_I$. The t-SNE processed matrix containing the features of all layers across all images is represented as:
\begin{equation}
	F^{N_I\times32\times1}=\text{t-SNE}(F^{N_I\times32\times4096}).
\end{equation}

\begin{figure}[htbp]
	\centering
	\includegraphics[width=.45\textwidth]{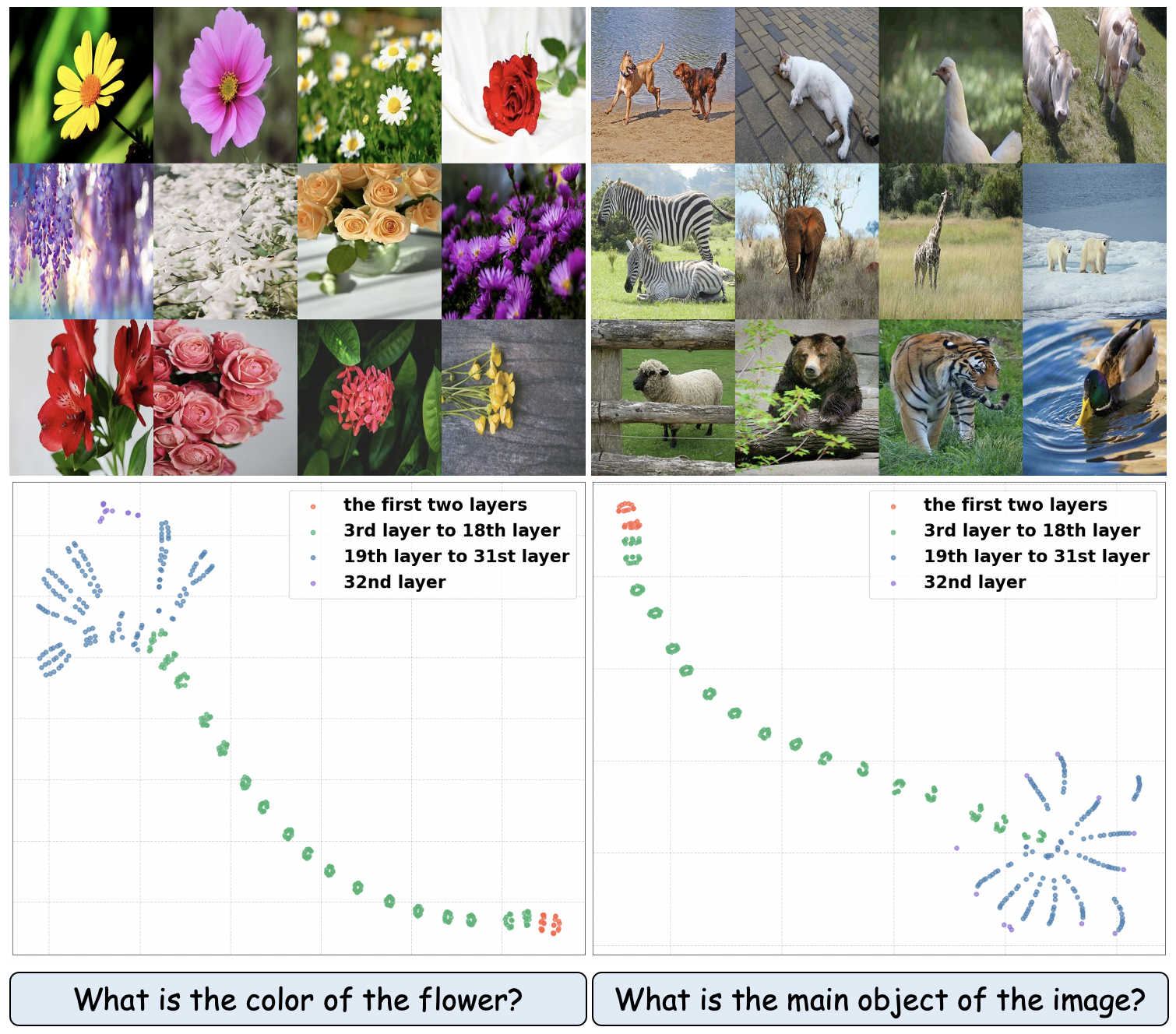}
	\caption{Feature encodings across layers for different images.}
	\label{fig:different}
\end{figure}

Two examples are shown in Figure \ref{fig:different}, each being observed on 12 different images. It can be seen that the feature encodings after t-SNE processing from different images, form an intriguing guitar-like shape: shallow features are closely clustered, forming the ``neck'' of the ``guitar'', while deep features diverge in different directions, forming the ``body'' of the ``guitar''. This indicates that in the network’s shallow layers, features extracted from different images are quite similar, but as the layers deepen, these features evolve to exhibit greater difference. The boundary between these shallow and deep layers coincides with what we previously identify as the critical layers. So far, we have consistently observed the critical layers across three aspects: for individual token probabilities, the critical layers mark the boundary where the probability of dominant tokens begins to rise sharply from near zero; for token distribution, they are nodes where intermediate information mutates from fluctuating to stabilizing; and for feature encoding, they are where features of different images transition from clustering to divergence. Additionally, we observe that the final layer’s features exhibit a distinct behavior compared to other layers, with some features showing a tendency to re-cluster. This is significantly evident in the left column, which represents tasks describing colors. A possible reason is that the attributes of the predicted tokens are relatively similar (e.g., animals, colors), which leads to similar encodings in the feature space, especially for the final transformer block used for generating the output tokens.
\section{Further Discussions}
\label{sec:discussion}

\paragraph{LVLM vs. LLM} Although all experiments in this paper are conducted on the LLM module of an LVLM, it is important to emphasize that the phenomena we observe are specific to LVLMs. For instance, when we ask LLaMA-1.5 ``Who was the first Indian to win an academic award?'' and examine the token distributions across layers, we find no clear evidence of hierarchy or mutations. For functional words, the distribution changes are minimal, while for significant tokens like names and dates, the distribution changes are more pronounced and continue through to the output layer. This suggests that the knowledge evolution process in LLMs differs significantly between text-only and multimodal inputs. This may be because LVLMs require instruction finetuning using vision-language data, which demands strict faithfulness to the input image, resulting in characteristics distinct from those of LLMs.
\vspace{-0.3cm}
\paragraph{Knowledge Evolution} Through the experiments in Section \ref{sec:token} - Section \ref{sec:feature}, we can summarize the general patterns of multimodal knowledge evolution in LVLMs: with the critical layers and mutation layers as two key nodes, the evolution process can be divided into three stages: rapid evolution, stabilization, and mutation. The rapid evolution stage occurs in the shallow layers of the model, where knowledge undergoes swift updates during forward propagation. As the layers deepen, the process transitions to the stabilization stage, where knowledge exhibits characteristics specific to tokens, and evolution slows. In the mutation stage, external knowledge not derived from the input image (e.g., general priors obtained through the LLM’s pretraining) is injected, prompting a secondary evolution. This newly introduced knowledge likely persists through to the output layer, ultimately influencing the selection of predicted tokens.
\vspace{-0.3cm}
\paragraph{Potential Applications} Our work analyzes the mechanism of knowledge evolution in LVLMs, which can aid researchers in designing more efficient and accurate models. We demonstrate that skipping some layers in the stabilization stage has minimal impact on the output. This insight supports the design of compression algorithms for lightweight models. Additionally, when the mutation layers are skipped, some hallucination tokens can be corrected, which is valuable for developing more effective hallucination mitigation algorithms. Furthermore, in Section \ref{sec:feature}, we observe that features cluster in shallow layers with different images. This observation aids in designing more efficient finetuning strategies: by finetuning only the deep-layer parameters, the model can generalize to new tasks.
\section{Conclusion}
\label{sec:conclusion}

We investigate how knowledge evolves within LVLMs. Using early exit and dimensionality reduction, we design several strategies for tracking knowledge, and explore towards this topic in three levels: single token probabilities, token probability distributions, and feature encodings. We provide the first insight into the process of knowledge evolution in LVLMs, based on two key nodes named as critical layers and mutation layers. Furthermore, based on the characteristics of different evolution stages, we explore novel perspectives on issues including model compression and hallucination elimination.
\section{Acknowledgements}
\label{sec:ack}

This work was supported in part by National Natural Science Foundation of China under grant No.62371450.
{
	\small
	\bibliographystyle{ieeenat_fullname}
	\bibliography{main}
}


\end{document}